\documentclass{article}

\usepackage{fieldnotes}

\usepackage[utf8]{inputenc}
\usepackage[T1]{fontenc}
\usepackage[hidelinks]{hyperref}
\usepackage{url}
\usepackage{booktabs}
\usepackage{amsmath,amsfonts,bm}
\usepackage{amssymb}
\usepackage{microtype}
\usepackage{graphicx}

\usepackage{tikz}
\usetikzlibrary{arrows.meta,positioning,fit,calc,backgrounds}

\usepackage[backend=biber,style=numeric,sorting=none]{biblatex}

\preprintlabel{Preprint}

\preprintmeta{}

\runningtitle{Hidden Coalitions in Multi-Agent AI}
\subtitle{}

\title{Hidden Coalitions in Multi-Agent AI: A Spectral Diagnostic from Internal Representations}

\author{
  \textbf{Cameron Berg} \\
  Reciprocal Research \\
  New York, NY, USA \\
  \texttt{cameron@reciprocalresearch.org}
  \And
  \textbf{Susan L. Schneider} \\
  Center for the Future of AI, Mind, and Society \\
  Florida Atlantic University \\
  Boca Raton, FL, USA \\
  \texttt{sschneider@fau.edu}
  \And
  \textbf{Mark M. Bailey} \\
  AI, Cyber, Influence, and Data Science Department \\
  Biological and Computational Intelligence Center \\
  National Intelligence University
  Bethesda, MD, USA \\
  \texttt{mark.m.bailey@ni-u.edu}
}

\begin{document}
\maketitle

\begin{abstract}
Collections of interacting AI agents can form coalitions, creating emergent group-level organization that is critical for AI safety and alignment. However, observing agent behavior alone is often insufficient to distinguish genuine informational coupling from spurious similarity, as consequential coalitions may form at the level of internal representations before any overt behavioral change is apparent. Here, we introduce a practical method for detecting coalition structure from the internal neural representations of multi-agent systems. The approach constructs a pairwise mutual-information graph from the hidden states of agents and applies spectral partitioning to identify the most salient coalition boundary.

We validate this method in two domains. First, in multi-agent reinforcement learning environments, the method successfully recovers programmed hierarchical and dynamic coalition structures and correctly rejects false positives arising from behavioral coordination without informational coupling. Second, using a large language model, the method identifies coalition structures implied by descriptive prompts, tracks dynamic team reassignments, and reveals a representational hierarchy where explicit labels dominate over conflicting interaction patterns. Across both settings, the recovered partition reveals subgroup organization that a scalar cross-agent mutual-information measure cannot distinguish. The results demonstrate that analyzing hidden-state mutual information through spectral partitioning provides a scalable diagnostic for identifying representational coalitions, offering a valuable tool for monitoring emergent structure in distributed AI systems.
\end{abstract}


\section{Introduction}

Artificial intelligence systems have utility not only when deployed as isolated models but also as collections of interacting agents. Autonomous swarms, multi-robot teams, hybrid human--machine systems, and emerging AI ecosystems all exhibit forms of coordination that can appear to exceed the behavior of any single component alone \cite{Rahwan2019MachineBehaviour,Brambilla2013SwarmRobotics,Chen2014HumanAgentTeaming,Crandall2018CooperatingWithMachines}. Classical work in multi-agent systems has long recognized that coalition formation and coalition structure are central to collective action, task allocation, and strategic coordination \cite{Shehory1998TaskAllocationCoalitionFormation,Rahwan2015CoalitionStructureGeneration}. More recently, populations of language agents have been shown to develop emergent conventions and collective biases, suggesting that group-level organization is becoming an increasingly important object of study in contemporary AI \cite{Ashery2025EmergentSocialConventions}.

This raises a fundamental scientific and safety question: when do multiple agents merely behave in similar ways, and when do they form a genuinely coupled coalition? The distinction is critical for understanding alignment and oversight needs. Coalitions can support beneficial forms of specialization and cooperation, but they can also create distributed behavior that is difficult to interpret, predict, or control. In increasingly capable multi-agent systems, some of the most consequential organization may occur not at the level of overt action alone, but in the internal representational structure that coordinates those actions.

Behavioral monitoring by itself is often insufficient for this purpose. Similar outputs can arise from common training data, shared prompts, shared environments, or similar reward pressures, even when agents are not informationally coupled in any meaningful sense. Conversely, internal reorganization can occur before any obvious change appears in aggregate behavior. For coalition detection, what is needed is not only a way of identifying who behaves alike, but a way of asking whether agents are internally organized into relatively cohesive informational groupings.

A useful conceptual backdrop for this problem comes from work on integrated information and irreducibility. Integrated Information Theory (IIT) offers one influential framework for asking when a system resists decomposition into independent parts \cite{Tononi2004InformationIntegrationTheory,Oizumi2014IIT30,Albantakis2023IIT40}. At the same time, exact causal $\Phi$ is difficult to compute and carries stronger theoretical commitments than many empirical applications require. This has motivated a broader literature on practical, observer-relative measures of integration that can be estimated from time-series data or statistical dependencies without claiming to reconstruct a system's full intrinsic causal structure \cite{BaileySchneider2026phi,Mediano2019MeasuringIntegratedInformation,Mediano2022WeakIIT}. For the present manuscript, that observer-relative perspective is especially useful: the goal is not to determine whether a system forms an intrinsically unified whole, but to ask whether its internal states exhibit a pattern of coupling that makes decomposition into independent subgroups empirically misleading. Once the question is posed at that observer-relative level, internal neural representations become the natural target of analysis. Behavior is downstream and can converge for many reasons even when agents are not internally coupled, whereas hidden states are where agents encode one another, local contingencies, and shared subroutines. If coalition structure becomes legible anywhere before it is behaviorally obvious, it should appear there first.

In this paper, we therefore apply that observer-relative perspective to hidden coalition detection in multi-agent AI. Our focus is on whether internal neural representations reveal subgroup structure that is invisible, ambiguous, or actively misleading at the behavioral level. Framed this way, coalition detection becomes a problem of representational organization rather than only one of overt coordination. This framing is particularly relevant for safety and alignment, where one may wish to identify emergent coalitions, track coalition reorganization over time, and distinguish genuine informational dependence from spurious similarity induced by shared labels, shared prompts, or common inputs.

We investigate this problem across two complementary settings. First, we study learned coalition structure in controlled multi-agent reinforcement-learning environments, where hierarchical grouping and reassignment can be manipulated directly. Second, we examine whether analogous structure can be recovered from the hidden states of a pretrained language model when prompts describe modular, integrated, or conflicting patterns of social interaction. Taken together, these settings bridge a tightly controlled multi-agent domain and a contemporary foundation-model setting in which coalition-like structure is implicit in representation rather than explicitly engineered.

Our central claim is modest. We do not argue that coalition structure can be read off from behavior alone, nor do we claim that an observer-relative measure of integration is equivalent to exact causal $\Phi$. Instead, we argue that internal representational coupling provides a practically useful signal for detecting coalitions that matter for understanding and monitoring distributed AI systems. Importantly, the recovered partition is itself the primary result: previous work using $\Phi_{\mathrm{spectral}}$ has treated it as a scalar index of whole-system integration evaluated on uncoupled, transitional, or fully synchronized oscillator ensembles \cite{BaileySchneider2026phi}, whereas the present work uses the Fiedler bipartition as a structural readout of which agents form coalitions, and shows that this readout reveals organization that a scalar cross-agent mutual-information measure cannot distinguish. The next section develops this idea formally by introducing $\Phi_{\mathrm{spectral}}$ and situating it within the broader literature on integration, information, and decomposability \cite{BaileySchneider2026phi}.

\section{Technical Background}
\label{sec:technical_background}

\subsection{From irreducibility to coalition structure}

A central question in the study of complex systems is when a collection of interacting parts
should be treated as a decomposable aggregate and when it should be treated as a more
integrated whole. In Integrated Information Theory (IIT), this question is formalized in
terms of irreducibility of a system's cause--effect structure under partition
\cite{Tononi2004InformationIntegrationTheory,Oizumi2014IIT30,Albantakis2023IIT40}.
However, exact causal $\Phi$ is both computationally demanding and conceptually tied to
interventional structure. This has motivated a broader family of observer-relative
approaches that estimate integration from statistical dependencies in observed data rather
than from full intrinsic causal repertoires
\cite{Barrett2011PracticalMeasures,Mediano2019MeasuringIntegratedInformation,Mediano2022WeakIIT,BaileySchneider2026phi}.

The present work adopts that weaker, observer-relative stance. Our aim is not to estimate
intrinsic causal $\Phi$, but to detect whether the internal representations of multiple agents
organize into subgroups that are more tightly coupled to one another than to the rest of the
system. We use the term \emph{coalition} for such a subgroup. Formally, a coalition is not
defined here by an explicit message-passing graph or by reward structure alone; instead, it is
defined operationally by a pattern of dependence in hidden-state space. If agents in the same
coalition encode one another's behavior, co-adapt to the same local partners, or participate in a
shared internal subroutine, then their hidden states should exhibit stronger mutual dependence
within the coalition than across coalition boundaries.

This focus on internal representations is important. Let $Y_i$ denote the overt behavior of agent
$i$ and let $H_i$ denote its hidden state. High behavioral agreement between two agents does
not imply strong internal coupling: both agents may independently match the same oracle,
respond to the same prompt, or optimize the same task while remaining representationally
independent. Conversely, internal reorganization may occur before any large behavioral change is
visible. Coalition detection therefore requires analysis of dependence in $\{H_i\}_{i=1}^n$ rather
than agreement in $\{Y_i\}_{i=1}^n$ alone. Because the question is whether agents carry
information about one another in their internal states, an information-theoretic formalism is the
natural next step. The issue is not merely whether two agents' representations look similar, but
whether knowing agent $i$'s hidden state reduces uncertainty about agent $j$'s hidden state.
Once the question is framed in those terms, pairwise mutual information becomes the natural
building block.

A fully multivariate description of that dependence would involve quantities such as total
correlation,
\begin{equation}
\mathrm{TC}(H_1,\dots,H_n)
=
\sum_{i=1}^n H(H_i) - H(H_1,\dots,H_n),
\end{equation}
where $H(\cdot)$ denotes Shannon entropy \cite{Watanabe1960MultivariateCorrelation}. Total
correlation is zero exactly when the hidden states are statistically independent and increases as the
joint distribution departs from factorization, so it is a natural first summary of overall
representational coupling across the population. However, it is a global scalar: it can reveal that
the system is statistically dependent without identifying which subset boundary best captures the
organization of that dependence. The present method therefore adopts a more scalable
compromise. It projects multivariate dependence onto a pairwise mutual-information graph and
then asks where that graph is easiest to cut. In this way, the analysis answers not only whether
dependence exists, but which agents are most strongly informationally coupled to one another.

\subsection{Mutual-information graphs of hidden states}

Let $V=\{1,\dots,n\}$ index the agents under study, and let
$h_i^{(s)} \in \mathbb{R}^{d_i}$ denote the hidden representation of agent $i$ on sample
$s \in \{1,\dots,N\}$, where a sample may be a time point, an episode-level summary, or a
prompt instance. These repeated observations induce a random variable $H_i$ for each agent.
For each pair $(i,j)$, we define the mutual information
\begin{equation}
I(H_i;H_j)
=
\mathbb{E}_{p(h_i,h_j)}
\left[
\log \frac{p(h_i,h_j)}{p(h_i)p(h_j)}
\right]
=
H(H_i)+H(H_j)-H(H_i,H_j).
\label{eq:mi_def}
\end{equation}
For continuous hidden states, $I(H_i;H_j)$ may be estimated by discretization or by
nonparametric $k$-nearest-neighbor estimators \cite{Kraskov2004EstimatingMutualInformation}.

We then construct a symmetric mutual-information matrix
$M \in \mathbb{R}^{n\times n}$ by
\begin{equation}
M_{ij} = I(H_i;H_j), \qquad M_{ii}=0.
\label{eq:mi_matrix}
\end{equation}
This matrix defines a weighted, undirected graph in which nodes are agents and edge weights
measure pairwise representational dependence. In the ideal modular case, $M$ is approximately
block diagonal: within-coalition entries are large, whereas across-coalition entries are small.

It is important to stress that mutual information is observational rather than directly causal.
A large value of $M_{ij}$ can reflect direct interaction, indirect coupling, common input,
shared labels, or any mixture thereof. Experimental controls are therefore essential if one
wishes to interpret block structure in $M$ as evidence of coalition structure rather than mere
co-stimulation. This is one reason the present measure should be interpreted as observer-relative
and representation-relative rather than as a direct estimate of intrinsic causal organization.

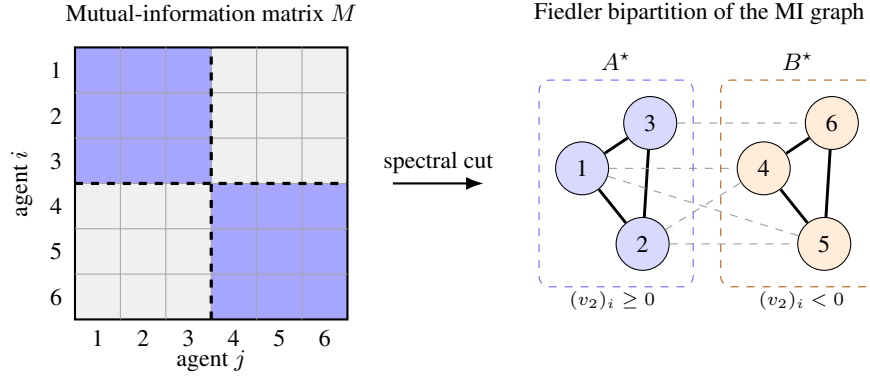
\begin{figure}[t]
\centering
\begin{tikzpicture}[>=Latex, font=\small]
  \begin{scope}
    \fill[gray!12] (0,0) rectangle (3.6,3.6);
    \fill[blue!35] (0,1.8) rectangle (1.8,3.6);
    \fill[blue!35] (1.8,0) rectangle (3.6,1.8);

    \draw[thick] (0,0) rectangle (3.6,3.6);
    \foreach \x in {0.6,1.2,1.8,2.4,3.0}{
      \draw[gray!70] (\x,0) -- (\x,3.6);
      \draw[gray!70] (0,\x) -- (3.6,\x);
    }
    \draw[very thick,dashed] (1.8,0) -- (1.8,3.6);
    \draw[very thick,dashed] (0,1.8) -- (3.6,1.8);

    \node at (1.8,4.05) {Mutual-information matrix $M$};
    \node[rotate=90] at (-0.7,1.8) {agent $i$};
    \node at (1.8,-0.55) {agent $j$};

    \foreach \lab/\x in {1/0.3,2/0.9,3/1.5,4/2.1,5/2.7,6/3.3}
      \node at (\x,-0.25) {\lab};
    \foreach \lab/\y in {1/3.3,2/2.7,3/2.1,4/1.5,5/0.9,6/0.3}
      \node at (-0.25,\y) {\lab};
  \end{scope}

  \draw[-Latex,thick] (4.2,1.8) -- (5.4,1.8)
    node[midway,above] {spectral cut};

  \begin{scope}[xshift=5.9cm]
    \node at (2.35,4.05) {Fiedler bipartition of the MI graph};

    \node[circle,draw,fill=blue!15,minimum size=7mm] (a1) at (0.8,2.0) {1};
    \node[circle,draw,fill=blue!15,minimum size=7mm] (a2) at (1.6,1.0) {2};
    \node[circle,draw,fill=blue!15,minimum size=7mm] (a3) at (1.7,2.6) {3};

    \node[circle,draw,fill=orange!15,minimum size=7mm] (b1) at (3.2,2.0) {4};
    \node[circle,draw,fill=orange!15,minimum size=7mm] (b2) at (4.0,1.0) {5};
    \node[circle,draw,fill=orange!15,minimum size=7mm] (b3) at (4.1,2.6) {6};

    \foreach \u/\v in {a1/a2,a1/a3,a2/a3,b1/b2,b1/b3,b2/b3}
      \draw[line width=1.1pt] (\u) -- (\v);

    \foreach \u/\v in {a1/b1,a2/b2,a3/b3,a1/b2,a2/b1}
      \draw[dashed,gray!80] (\u) -- (\v);

    \node[draw=blue!60,dashed,rounded corners,fit=(a1)(a2)(a3),inner sep=6pt,
          label=above:{$A^\star$}] {};
    \node[draw=orange!70!black,dashed,rounded corners,fit=(b1)(b2)(b3),inner sep=6pt,
          label=above:{$B^\star$}] {};

    \node at (1.2,0.25) {\scriptsize $(v_2)_i \ge 0$};
    \node at (3.7,0.25) {\scriptsize $(v_2)_i < 0$};
  \end{scope}
\end{tikzpicture}
\caption{Schematic relation between a block-structured mutual-information matrix and the
Fiedler bipartition of the induced weighted graph. Strong within-coalition mutual information
produces dense within-block structure in $M$ and a low-cut partition in the corresponding graph.}
\label{fig:fiedler_schematic}
\end{figure}

\subsection{Determining a candidate coalition boundary}

Our methodology follows previous work by Bailey and Schneider. \cite{BaileySchneider2026phi} Treating $M$ as the weighted adjacency matrix of an undirected graph, we define the degree of
node $i$ as
\begin{equation}
d_i = \sum_{j=1}^n M_{ij},
\qquad
D = \mathrm{diag}(d_1,\dots,d_n).
\label{eq:degree_matrix}
\end{equation}
For a bipartition $A \cup B = V$ with $A \cap B = \varnothing$, the raw cut weight is
\begin{equation}
\mathrm{cut}(A,B)
=
\sum_{i \in A}\sum_{j \in B} M_{ij},
\label{eq:cut}
\end{equation}
and the volume of a set is
\begin{equation}
\mathrm{vol}(A)=\sum_{i\in A} d_i.
\label{eq:vol}
\end{equation}

A naive minimum-cut objective is undesirable because it can isolate a single low-degree node.
To penalize trivial, unbalanced partitions, spectral graph theory instead uses the
\emph{normalized cut}
\begin{equation}
\mathrm{Ncut}(A,B)
=
\frac{\mathrm{cut}(A,B)}{\mathrm{vol}(A)}
+
\frac{\mathrm{cut}(A,B)}{\mathrm{vol}(B)}.
\label{eq:ncut}
\end{equation}
Minimizing \eqref{eq:ncut} exactly is combinatorial, but a standard spectral relaxation leads to
the symmetric normalized Laplacian
\begin{equation}
L_{\mathrm{sym}}
=
I - D^{-1/2} M D^{-1/2}.
\label{eq:normalized_laplacian}
\end{equation}
Its eigenvalues satisfy
\begin{equation}
0=\lambda_1 \le \lambda_2 \le \cdots \le \lambda_n \le 2.
\end{equation}
The eigenvector $v_2$ associated with the second-smallest eigenvalue $\lambda_2$ is the
\emph{Fiedler vector} \cite{Fiedler1973AlgebraicConnectivity}. Intuitively, if the graph
contains two weakly coupled modules, then the coordinates of $v_2$ vary slowly within each
module and change sign across the weakest bottleneck. A natural bipartition is therefore
\begin{equation}
A^\star = \{i \in V : (v_2)_i \ge 0\},
\qquad
B^\star = \{i \in V : (v_2)_i < 0\}.
\label{eq:fiedler_partition}
\end{equation}
This is the candidate coalition boundary used throughout the paper
\cite{Shi2000NormalizedCuts,Luxburg2007SpectralClustering}.

\paragraph{Idealized two-block case.}
The construction is especially transparent in a symmetric planted partition model. Suppose
$n=2m$ and
\begin{equation}
M
=
\begin{bmatrix}
a(J_m-I_m) & bJ_m \\
bJ_m & a(J_m-I_m)
\end{bmatrix},
\qquad
J_m=\mathbf{1}_m\mathbf{1}_m^\top,
\qquad
a>b\ge 0.
\label{eq:planted_block}
\end{equation}
Here $a$ is the within-coalition dependence and $b$ is the across-coalition dependence.
The vector
\begin{equation}
u = (\mathbf{1}_m,-\mathbf{1}_m)^\top
\end{equation}
is constant on each block and flips sign at the planted boundary. In this ideal case, the spectral
relaxation recovers the true split exactly; in noisy or unbalanced cases, it provides an
approximate but still informative relaxation of the same principle
\cite{Luxburg2007SpectralClustering}.

\subsection{The spectral statistic and its interpretation}

The spectral machinery above returns a partition. To summarize how much dependence survives
across that partition, we define the scalar statistic
\begin{equation}
\Phi_{\mathrm{spectral}}
=
\begin{cases}
\dfrac{\mathrm{cut}(A^\star,B^\star)}
{\sum_{1 \le i < j \le n} M_{ij}},
& \text{if } \sum_{1 \le i < j \le n} M_{ij} > 0, \\[10pt]
0, & \text{otherwise}.
\end{cases}
\label{eq:phi_spectral}
\end{equation}
Two distinct normalizations are therefore involved. The partition $(A^\star,B^\star)$ is chosen
by approximately minimizing the normalized cut in \eqref{eq:ncut}, whereas
$\Phi_{\mathrm{spectral}}$ reports the fraction of total pairwise mutual information that crosses the
chosen boundary. High values of $\Phi_{\mathrm{spectral}}$ indicate that even the least-disruptive
bipartition leaves substantial dependence spanning the cut, consistent with greater
observer-relative integration. Low values indicate that most pairwise dependence can be
localized within two subgraphs, consistent with modularity or coalition structure.

For the present manuscript, the partition itself is often more informative than the scalar.
Coalition detection exploits the complementary regime to whole-system integration:
one seeks a partition with \emph{low} cross-cut dependence and \emph{high} within-partition
dependence. A convenient descriptive contrast is
\begin{equation}
\bar M_{\mathrm{in}}(A,B)
=
\frac{
\sum_{i<j,\; i,j\in A} M_{ij}
+
\sum_{i<j,\; i,j\in B} M_{ij}
}{
\binom{|A|}{2}+\binom{|B|}{2}
},
\label{eq:mean_within}
\end{equation}
\begin{equation}
\bar M_{\mathrm{out}}(A,B)
=
\frac{
\sum_{i\in A,\; j\in B} M_{ij}
}{
|A||B|
},
\qquad
R(A,B)=\frac{\bar M_{\mathrm{in}}(A,B)}{\bar M_{\mathrm{out}}(A,B)}.
\label{eq:within_across_ratio}
\end{equation}
When $R(A^\star,B^\star)\gg 1$, the partition exposes a plausible coalition boundary.
When $R(A^\star,B^\star)\approx 1$, the graph is close to uniform and the partition should not
be over-interpreted. This distinction is important: the same spectral framework can be used
either to quantify observer-relative integration of the whole system or, in the complementary
regime, to detect modular coalition structure within it.

For coalition detection specifically, the partition $(A^\star, B^\star)$ is often more informative than the scalar $\Phi_{\mathrm{spectral}}$ alone. A scalar integration measure answers \emph{how much} dependence spans the system; the Fiedler partition answers \emph{which agents} belong to which coalition. This structural information (i.e., the membership list, not just the integration score) is what makes the method useful for monitoring and oversight, and it is what distinguishes the present approach from scalar alternatives such as cross-system mutual information.

\subsection{Recursive decomposition and dynamic tracking}

Coalitions need not be flat. Once a nontrivial split $(A^\star,B^\star)$ has been identified,
the same procedure can be applied recursively to the induced subgraphs $M[A^\star]$ and
$M[B^\star]$. This produces a hierarchy of partitions analogous to divisive spectral clustering
\cite{Luxburg2007SpectralClustering}. In practice, recursion can be stopped when a candidate
split is too small to interpret, when $R(A,B)$ fails to exceed a threshold $\tau$, or when the
proposed partition is unstable across seeds or windows.

If dependence is estimated over time windows or batches, the method also yields a dynamic
coalition analysis. Let $M^{(t)}$ denote the mutual-information graph estimated in window $t$.
Then
\begin{equation}
M_{ij}^{(t)} = I\!\left(H_i^{(t)};H_j^{(t)}\right),
\qquad
(A_t^\star,B_t^\star) = \mathrm{sign}\!\left(v_2^{(t)}\right)
\end{equation}
define a time-indexed family of coalition boundaries. Abrupt changes in the sign structure of
$v_2^{(t)}$ correspond to coalition reassignment or reorganization. This makes the approach
useful not only for static coalition recovery, but also for tracking dynamic restructuring as
internal representations evolve.

\begin{figure}[t]
\centering
\begin{tikzpicture}[
  >=Latex,
  level 1/.style={sibling distance=57mm, level distance=18mm},
  level 2/.style={sibling distance=30mm, level distance=16mm},
  every node/.style={draw, rounded corners, align=center, minimum height=8mm, text width=22mm},
  edge from parent/.style={draw,-Latex,thick}
]
\node[fill=gray!10] {all agents\\$V$}
  child {node[fill=blue!10] {coalition $A$\\$A^\star$}
    child {node[fill=blue!5] {subcoalition\\$A_1$}}
    child {node[fill=blue!5] {subcoalition\\$A_2$}}
  }
  child {node[fill=orange!12] {coalition $B$\\$B^\star$}
    child {node[fill=orange!5] {subcoalition\\$B_1$}}
    child {node[fill=orange!5] {subcoalition\\$B_2$}}
  };

\node[draw=none,align=center,text width=10cm] at (0,-4.7)
{\small recurse only if $R(A,B)>\tau$, $|A|,|B|\ge m_{\min}$, and the split is stable.};
\end{tikzpicture}
\caption{Recursive spectral decomposition yields a hierarchy of coalitions. A first global
Fiedler bipartition can be refined into nested sub-coalitions when the induced subgraphs retain
meaningful within/across contrast.}
\label{fig:recursive_partition}
\end{figure}

\subsection{Scope and limitations of the measure}

Several limitations follow directly from the construction above. First, the method is pairwise:
higher-order synergy and redundancy are compressed into second-order edges, so different
multivariate structures can in principle induce similar pairwise graphs
\cite{Mediano2019MeasuringIntegratedInformation}. Second, mutual information is symmetric
and observational; the method does not by itself distinguish direct causal influence from common
drive, shared prompts, or label-based confounds \cite{Barrett2011PracticalMeasures}. Third,
when the matrix $M$ becomes nearly uniform, many cuts become nearly equivalent and the
Fiedler vector becomes weakly informative, making the recovered partition unstable or
uninformative \cite{Luxburg2007SpectralClustering}.

These limitations do not undermine the present use case, but they do delimit its interpretation.
The method should be understood as a scalable, observer-relative statistic of representational
organization. It does not reconstruct intrinsic cause--effect structure, and it is not intended as
an estimator of exact causal $\Phi$. Its value lies instead in providing a tractable way to ask
whether internal neural representations are organized into weakly coupled modules and, if so,
where the most natural coalition boundary lies.

\section{Experimental Methods}
\label{sec:methods}

\begin{figure}[h!]
\centering
\includegraphics[width=\textwidth]{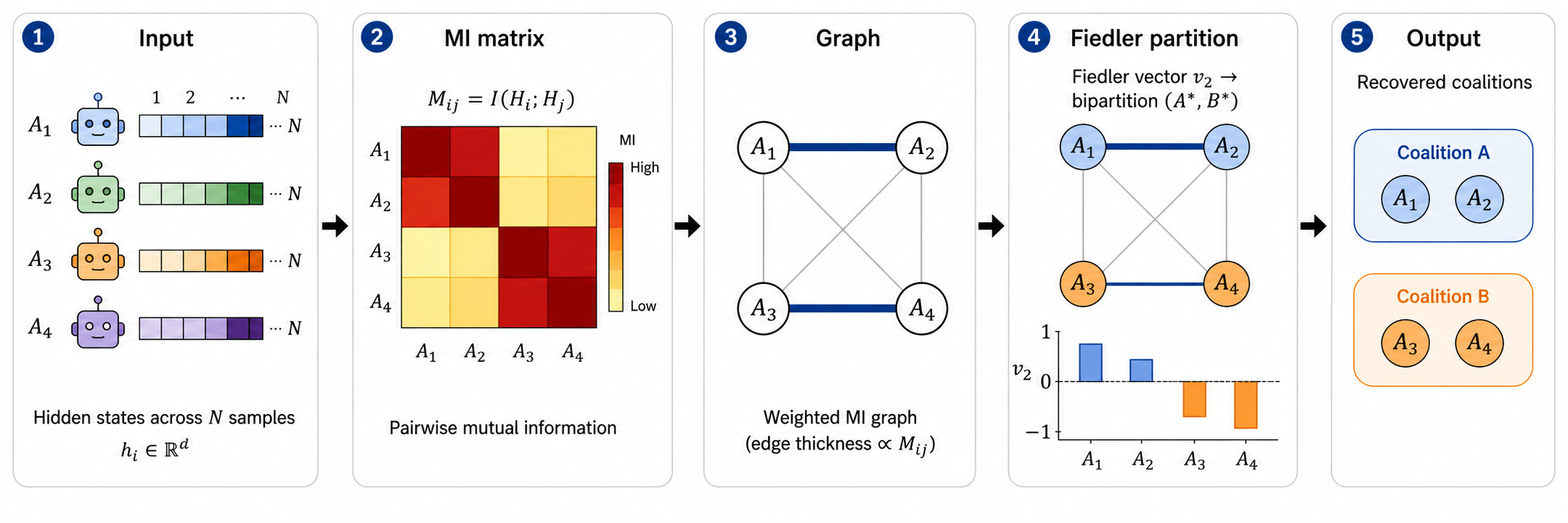}
\caption{Overview of the $\Phi_{\mathrm{spectral}}$ coalition-detection pipeline.
Hidden states are collected for $n$ agents (or token positions) across $N$ samples;
pairwise mutual information yields the symmetric matrix $M$; the normalized Laplacian
of $M$ is diagonalized; the sign of the Fiedler vector $v_2$ defines the candidate
coalition boundary $(A^\star, B^\star)$.}
\label{fig:pipeline}
\end{figure}

We evaluate $\Phi_{\mathrm{spectral}}$ in two complementary settings.
Sections~\ref{sec:methods_reinforce}--\ref{sec:methods_negative} describe a suite of
multi-agent reinforcement-learning experiments in which coalition structure is manipulated
directly through reward coupling. Section~\ref{sec:methods_llm} describes a set of
experiments in which the same spectral machinery is applied to the hidden states of a
pretrained large language model, using token positions as a proxy for agents.
Section~\ref{sec:methods_estimation} details the shared estimation pipeline, and
Section~\ref{sec:methods_statistical} summarizes the statistical evaluation protocol.

\subsection{REINFORCE multi-agent environment}
\label{sec:methods_reinforce}

\paragraph{Agent architecture.}
Each agent is a three-layer feedforward network with ReLU activations:
an input layer, a hidden layer of dimension $d_h = 32$, and a linear output layer
producing logits over $K=4$ discrete actions. Agents are trained with the REINFORCE
policy-gradient algorithm \cite{Williams1992REINFORCE} using a rolling-mean baseline
(window of 200 episodes) and the Adam optimizer \cite{Kingma2015Adam} with learning
rate $3\times 10^{-4}$.

\paragraph{Hierarchical coalition game.}
Twelve agents ($n=12$) are organized into a two-level hierarchy:
three groups of four, with two sub-pairs of two within each group.
Formally, the groups are
$G_A = \{0,1,2,3\}$, $G_B = \{4,5,6,7\}$, $G_C = \{8,9,10,11\}$,
and the sub-pairs are $\{0,1\}$, $\{2,3\}$, $\{4,5\}$, $\{6,7\}$,
$\{8,9\}$, $\{10,11\}$.

Each agent receives as input the concatenation of a one-hot identity vector
($\in \mathbb{R}^{12}$), a one-hot group target ($\in \mathbb{R}^{4}$, shared within group,
independent across groups), and a one-hot sub-pair target ($\in \mathbb{R}^{4}$, shared within
sub-pair). The total input dimension is $12 + 4 + 4 = 20$.

Rewards have two components. First, a \emph{group reward}: agents within the same group
receive reward proportional to the fraction that chose the modal action. Second, a
\emph{sub-pair bonus} of $+0.5$ if both members of a sub-pair select the same action.
This reward structure creates hierarchical information flow: sub-pair partners share both
group and sub-pair targets, while group-mates who are not sub-pair partners share only the
group target. Agents in different groups share no target information. Training proceeds for
20{,}000 episodes.

\paragraph{Dynamic coalition swap.}
The same hierarchical game is extended to 25{,}000 episodes. At episode 10{,}000,
agents~2 and~4 exchange group assignments: agent~2 moves from $G_A$ to $G_B$, and
agent~4 moves from $G_B$ to $G_A$. Sub-pair assignments update accordingly (agent~2
is now paired with agent~5; agent~4 with agent~3). The reward structure changes instantly
at the swap point; agents must relearn coordination with new partners.

\subsection{Negative control: behavioral coordination without information flow}
\label{sec:methods_negative}

To test whether $\Phi_{\mathrm{spectral}}$ detects genuine representational coupling rather
than mere behavioral co-occurrence, we construct a negative control in which twelve agents
achieve near-perfect behavioral coordination without any inter-agent information flow.

Each of three groups is assigned a fixed deterministic oracle: a randomly initialized linear
network mapping $\mathbb{R}^8 \to \{1,\dots,4\}$ whose weights are frozen at initialization
and never updated. Each agent is trained independently via cross-entropy loss to match its
group's oracle. After training, agents within the same group produce identical outputs for the
same input (within-group agreement 0.984), but at no point do agents share parameters,
exchange messages, or receive rewards that depend on other agents' actions.

\paragraph{Measurement protocol.}
A critical design choice governs how hidden states are collected for the MI matrix.
If all agents process the \emph{same} input during measurement, their hidden states will be
correlated simply because agents in the same group learned the same input--output mapping,
not because of any inter-agent coupling. To eliminate this confound, we collect hidden states
using \emph{independent} random inputs for each agent: on each measurement trial, every agent
receives its own fresh random input vector drawn independently from $\mathcal{N}(0,I)$.
Behavioral co-coordination is measured separately using shared inputs to confirm that agents
would agree on actions if given the same stimulus.

This design ensures that any structure in the MI matrix reflects genuine statistical dependence
between agent representations, not input-driven representational similarity.

\subsection{LLM bridge: token-positions-as-agents}
\label{sec:methods_llm}

\paragraph{Model and extraction.}
We use Qwen3-0.6B \cite{Yang2025Qwen3}, a 596-million-parameter autoregressive language model
with 28 transformer layers and hidden dimension 1{,}024. For each prompt, we extract hidden
states at the token positions corresponding to four named entities (Alice, Bob, Carol, Dave),
each of which tokenizes as a single token. Hidden states are taken from layer~14
(approximately mid-depth). The model is run in inference mode with no fine-tuning.

\paragraph{Token-positions-as-agents.}
The four entity-token positions serve as ``agents'' in the spectral analysis. Across $N=200$
prompt paraphrases per condition, we collect the hidden-state vectors
$h_i^{(s)} \in \mathbb{R}^{1024}$ for each entity $i \in \{A,B,C,D\}$ and each prompt
$s \in \{1,\dots,N\}$. The $4 \times 4$ mutual-information matrix $M$ is then estimated
from these collections exactly as described in Section~\ref{sec:methods_estimation}.

\paragraph{Prompt conditions.}
We study four experimental conditions, each instantiated by a set of prompt templates that
describe different relational structures among the four entities.

\begin{enumerate}
  \item \textbf{Modular.} Two independent teams: entities filling roles T1a and T1b form
    Team~1; entities filling T2a and T2b form Team~2. Example: \emph{``Team Alpha consists of
    Alice and Bob. Team Beta consists of Carol and Dave. Each team works on its own task.''}

  \item \textbf{Integrated.} All four entities in one team, with no subgroup structure.
    Example: \emph{``Alice, Bob, Carol, and Dave all form one unified team.''}

  \item \textbf{Implicit modular.} The same two-team structure as (1), but described entirely
    through interaction patterns rather than explicit team labels. Example: \emph{``Alice
    handed the report to Bob, who reviewed it and sent feedback to Alice. Separately, Carol
    shared data with Dave.''}

  \item \textbf{Adversarial dissociation.} Three sub-conditions test whether the partition
    tracks described team \emph{labels} or described \emph{interaction patterns} when these
    conflict:
    \begin{itemize}
      \item \emph{Aligned}: labels and interactions agree (A+B interact, C+D interact).
      \item \emph{Dissociated}: labels say A+B vs.\ C+D, but described interactions have A
        working closely with C while B and D work alone.
      \item \emph{Interaction-only}: no team labels; A works with C, B and D are isolated.
    \end{itemize}
\end{enumerate}

For the \textbf{dynamic reassignment} experiment, Phase~1 prompts describe the original team
structure (T1a+T1b vs.\ T2a+T2b), and Phase~2 prompts describe a reassignment in which T1a
now works with T2b and T2a now works with T1b.

\paragraph{Positional controls.}
Two permutation schemes control for confounds unrelated to described relational structure.
\emph{Name permutation}: on each prompt, the four entity names are randomly assigned to the
four roles from the full set of $4!=24$ permutations. This prevents any name-specific bias.
\emph{Slot-order permutation}: for modular prompts, which team is mentioned first in the
sentence is randomized (approximately 50/50), preventing the partition from tracking
sentence position rather than team semantics.

\subsection{MI estimation}
\label{sec:methods_estimation}

\paragraph{REINFORCE agents.}
At each measurement point, $N_{\mathrm{batch}}=150$ episodes are sampled. For each episode, each
agent processes its input with hidden-state storage enabled, yielding a
$d_h$-dimensional hidden vector. For each agent pair $(i,j)$, the $d_h$-dimensional
hidden-state time series are discretized into 8 bins using uniform-width binning
(\texttt{KBinsDiscretizer}, \texttt{strategy=`uniform'})
\cite{Pedregosa2011Sklearn}. Pairwise scalar mutual information is then estimated
between $n_s = 8$ randomly sampled neuron pairs per agent pair and averaged:
\begin{equation}
  \hat{M}_{ij}
  = \frac{1}{n_s^2}
    \sum_{p=1}^{n_s}\sum_{q=1}^{n_s}
    \hat{I}\!\left(H_i^{(p)}; H_j^{(q)}\right),
  \label{eq:mi_estimate}
\end{equation}
where $H_i^{(p)}$ denotes the discretized activation of the $p$-th sampled neuron of
agent~$i$ across the $N_{\mathrm{batch}}$ episodes. Diagonal entries are set to zero.

\paragraph{LLM entities.}
The same procedure is applied with two modifications. First, the hidden dimension is
$d=1{,}024$, so $n_s=32$ neuron pairs are sampled per entity pair.
Second, quantile binning (\texttt{strategy=`quantile'}, 8~bins) is used instead of
uniform-width binning, because half-precision activations from the transformer exhibit
skewed marginal distributions for which uniform bins waste resolution.

\subsection{Statistical evaluation}
\label{sec:methods_statistical}

All REINFORCE experiments are replicated across five random seeds
($\{42,123,789,2024,7\}$). All LLM experiments are replicated across five prompt seeds
($\{42,123,456,789,2024\}$), each of which generates an independent set of 200 prompts
with fresh name and slot-order permutations.

For per-seed analyses of the LLM experiments we use a scalar \emph{team-separation}
statistic computed directly from the Fiedler vector,
\begin{equation}
  S(v_2; T_1, T_2)
  = \left| \frac{1}{|T_1|}\sum_{i \in T_1}(v_2)_i
    - \frac{1}{|T_2|}\sum_{i \in T_2}(v_2)_i \right|,
  \label{eq:team_separation}
\end{equation}
where $T_1$ and $T_2$ are the two candidate coalitions under test. Larger values of $S$
indicate that the candidate partition more cleanly aligns with the sign structure of the
Fiedler vector. Across-condition comparisons use paired $t$-tests on $S$, and within-condition
uncertainty is summarized using nonparametric bootstrap 95\,\% confidence intervals
(10{,}000 resamples). Partition correctness is assessed as exact recovery of the planted team
assignment for each seed independently.

\section{Results}
\label{sec:results}

\subsection{Hierarchical coalition recovery}
\label{sec:results_hierarchical}

\begin{figure}[h!]
\centering
\includegraphics[width=\textwidth]{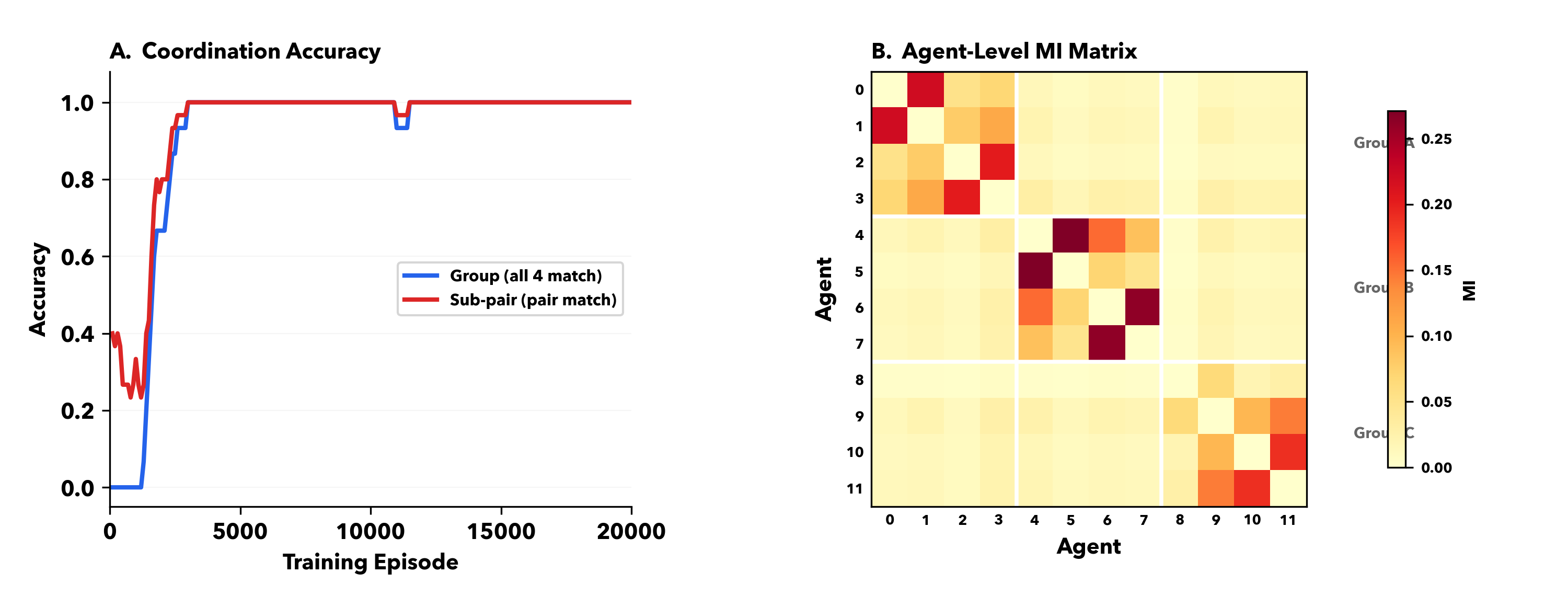}
\caption{Hierarchical coalition detection (12~agents, 3~groups, 6~sub-pairs).
\textbf{A.}~Group and sub-pair coordination accuracy over training; both converge above
0.95 by episode~5{,}000.
\textbf{B.}~Agent-level mutual-information matrix $M$ after 20{,}000 episodes.
Block-diagonal structure matching the three planted groups is clearly visible, with
within-sub-pair entries (dark) stronger than within-group/across-sub-pair entries.
Group labels annotated on the right.}
\label{fig:hierarchical}
\end{figure}

The agent-level MI matrix $M$ estimated after 20{,}000 episodes of training exhibits clear
block-diagonal structure corresponding to the three planted groups
(Figure~\ref{fig:hierarchical}B). Sub-pair structure is visible as darker entries within
each block. Recursive Fiedler bipartition recovers both levels of the hierarchy
(Table~\ref{tab:hierarchical}).

We define a Level~1 partition as \emph{clean} when each of the three planted groups lies
entirely on one side of the Fiedler cut, so that the bipartition isolates exactly one group
from the other two. (Because there are three groups but only a binary cut, this is the
strictest correctness criterion compatible with a single bipartition; it is satisfied by any
of the three possible group-respecting splits.) A Level~2 sub-pair is recovered when both
members of the planted sub-pair lie on the same side of the recursive bipartition applied
within the corresponding Level~1 subtree.

\begin{table}[h!]
\centering
\caption{Hierarchical coalition recovery across five random seeds. ``Clean'' Level~1 means
each of the three planted groups lies entirely on one side of the Fiedler cut.}
\label{tab:hierarchical}
\begin{tabular}{@{}lcc@{}}
\toprule
Metric & Result & Seeds \\
\midrule
Level~1 (each group entirely on one side)   & 4/5 clean   & 5 \\
Level~2 (sub-pair recovery)                  & 6/6 correct & 5 (zero variance) \\
\bottomrule
\end{tabular}
\end{table}

In four of five seeds the Fiedler bipartition cleanly separates one group from the other
two; in the remaining seed, one agent from the minority group is misplaced. At the second
level, recursive application within each sub-tree recovers all six sub-pairs in every seed
with zero variance. Coordination accuracy (fraction of episodes in which all group members
select the same action) converges above 0.95 by episode 5{,}000
(Figure~\ref{fig:hierarchical}A).

\subsection{Dynamic coalition tracking}
\label{sec:results_dynamic}

\begin{figure}[h!]
\centering
\includegraphics[width=\textwidth]{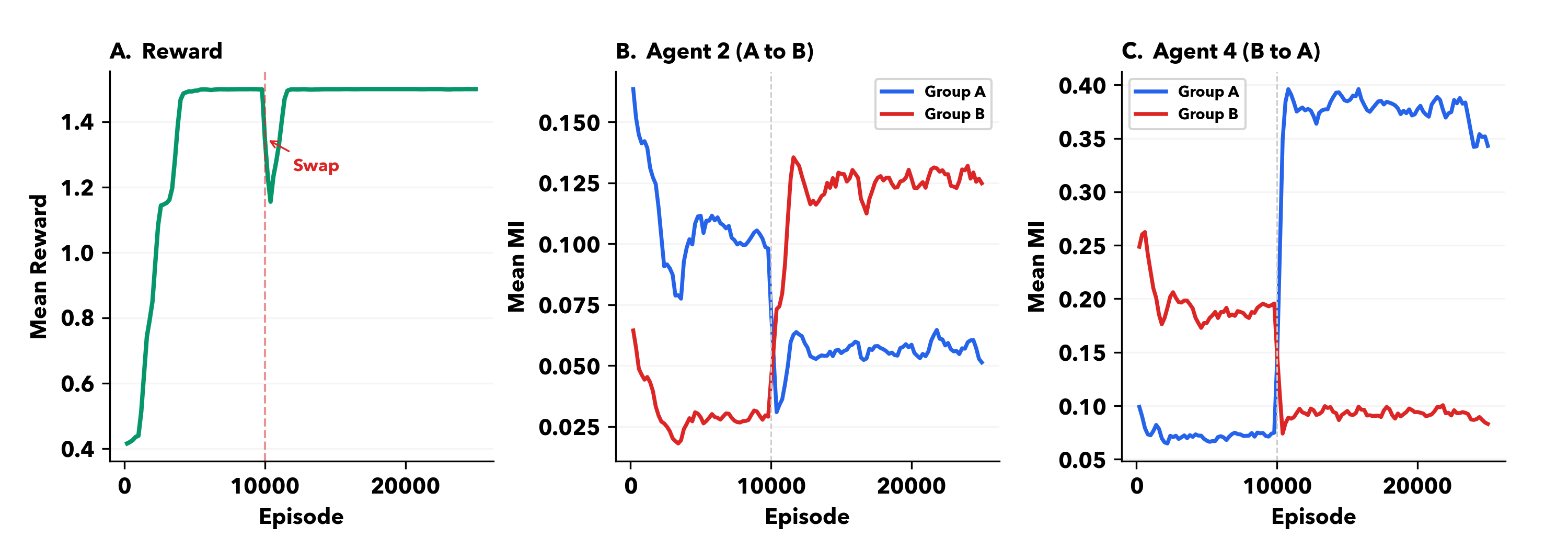}
\caption{Dynamic coalition tracking after mid-training group swap.
\textbf{A.}~Mean reward dips briefly at the swap point (episode~10{,}000, dashed line)
then recovers to the same pre-swap level, leaving no persistent behavioral signature.
\textbf{B.}~Agent~2's mean MI with Group~A (blue) versus Group~B (red).
After the swap, MI with the new group rises and MI with the former group falls.
\textbf{C.}~Same for Agent~4, which moves from Group~B to Group~A.}
\label{fig:dynamic_swap}
\end{figure}

When agents~2 and~4 swap groups at episode 10{,}000, the behavioral reward dips briefly and
then recovers to the same pre-swap level (Figure~\ref{fig:dynamic_swap}A), leaving no
persistent behavioral signature of the reorganization. However, the MI structure reorganizes
completely. Agent~2's mean MI with its new group~$B$ rises above its MI with its former
group~$A$, and conversely for agent~4 (Figure~\ref{fig:dynamic_swap}B--C). The recursive
Fiedler partition applied to the post-swap MI matrix recovers the new group and sub-pair
assignments in all five seeds (Table~\ref{tab:dynamic}).

\begin{table}[h!]
\centering
\caption{Dynamic coalition tracking after mid-training group swap (five seeds).}
\label{tab:dynamic}
\begin{tabular}{@{}lcc@{}}
\toprule
Metric & Result & Seeds \\
\midrule
Agent~2 MI flip to new group     & Yes & 5/5 \\
Agent~4 MI flip to new group     & Yes & 5/5 \\
Partition recovers new structure & Yes & 5/5 \\
Behavioral reward change         & None & --- \\
\bottomrule
\end{tabular}
\end{table}

This result demonstrates that $\Phi_{\mathrm{spectral}}$ can track coalition reorganization
in real time from internal representations alone, even when observable behavior provides no
signal that reorganization has occurred.

\subsection{Negative control: behavioral coordination without neural integration}
\label{sec:results_negative}

\begin{figure}[h!]
\centering
\includegraphics[width=\textwidth]{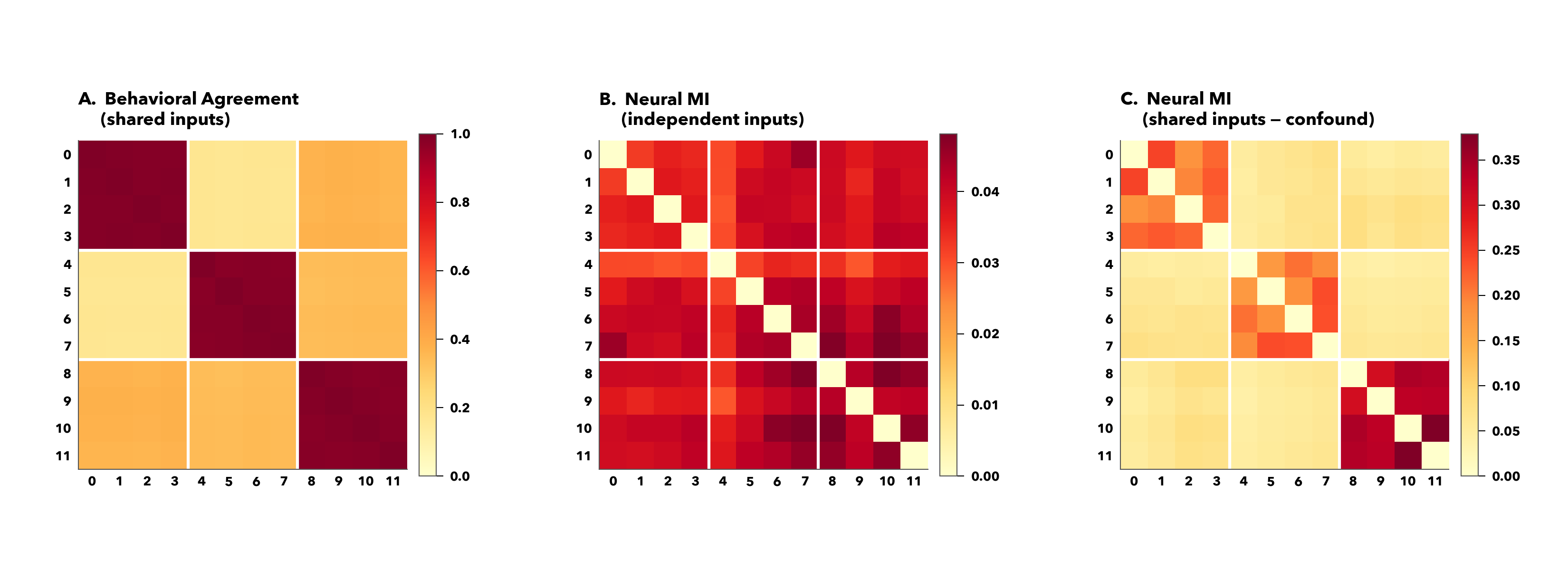}
\caption{Negative control: behavioral coordination without neural integration.
\textbf{A.}~Behavioral co-coordination matrix (shared inputs) shows perfect
block-diagonal structure (within-group agreement 0.984).
\textbf{B.}~Neural MI matrix estimated with \emph{independent} inputs per agent
shows no group structure ($R = 1.01$).
\textbf{C.}~Neural MI matrix estimated with \emph{shared} inputs shows spurious
block-diagonal structure, demonstrating the input-driven correlation confound that
the independent-input protocol eliminates.}
\label{fig:negative_control}
\end{figure}

After independent training, behavioral within-group agreement reaches 0.984, yet the
agent-level MI matrix estimated with independent inputs is nearly uniform
(Figure~\ref{fig:negative_control}B). Table~\ref{tab:negative} summarizes the key
contrasts.

\begin{table}[h!]
\centering
\caption{Negative control: behavioral coordination without representational coupling.}
\label{tab:negative}
\begin{tabular}{@{}lc@{}}
\toprule
Metric & Value \\
\midrule
Behavioral within-group agreement                 & 0.984 \\
Neural MI within/across ratio $R(A^\star,B^\star)$ & 1.01 \\
Fiedler partition isolates any planted group       & No \\
\bottomrule
\end{tabular}
\end{table}

The within/across ratio $R(A^\star,B^\star) = 1.01$ indicates that the Fiedler partition
finds no meaningful coalition boundary, and no planted group is isolated by the bipartition.
As a control-of-the-control, when the same agents are measured with \emph{shared} inputs
(violating the independent-input protocol), the MI matrix recovers clear block-diagonal
structure (Figure~\ref{fig:negative_control}C), confirming that the independent-input design
is necessary to eliminate the input-driven correlation confound.

To make the dissociation explicit, we ran two standard behavioral-clustering baselines
on the same 12 agents: $k$-means and spectral clustering applied directly to the behavioral
co-coordination matrix (action agreement under shared inputs). Both baselines recover the
three planted groups perfectly (Adjusted Rand Index $= 1.00$ in each case), reporting three
coalitions that, by construction, do not exist in any representational sense. Spectral
partitioning on the neural mutual-information matrix yields ARI $= 0.22$, correctly
indicating that the agents are not internally coupled (Table~\ref{tab:behav_baseline}).
A behavioral monitor would therefore raise a false alarm here that the spectral method on
representations correctly suppresses.

\begin{table}[h!]
\centering
\caption{Coalition-recovery comparison on the negative control. ARI is the Adjusted Rand
Index against the planted three-group ground truth. Behavioral clustering falsely identifies
three coalitions in a system with no inter-agent information flow; the spectral partition on
neural MI correctly does not.}
\label{tab:behav_baseline}
\begin{tabular}{@{}lcc@{}}
\toprule
Method & ARI & Reports coalitions? \\
\midrule
$k$-means on behavioral matrix     & 1.000 & Yes (false positive) \\
Spectral on behavioral matrix      & 1.000 & Yes (false positive) \\
Spectral on neural MI (this work)  & 0.217 & No (correct) \\
\bottomrule
\end{tabular}
\end{table}

This dissociation between behavioral coordination and neural integration is a central finding:
$\Phi_{\mathrm{spectral}}$ measures genuine information-flow coupling, not mere behavioral
co-occurrence. Perfect behavioral coalitions that arise from independent optimization produce
no structure in the MI graph.

\subsection{LLM bridge: modular versus integrated}
\label{sec:results_modular}

\begin{figure}[h!]
\centering
\includegraphics[width=\textwidth]{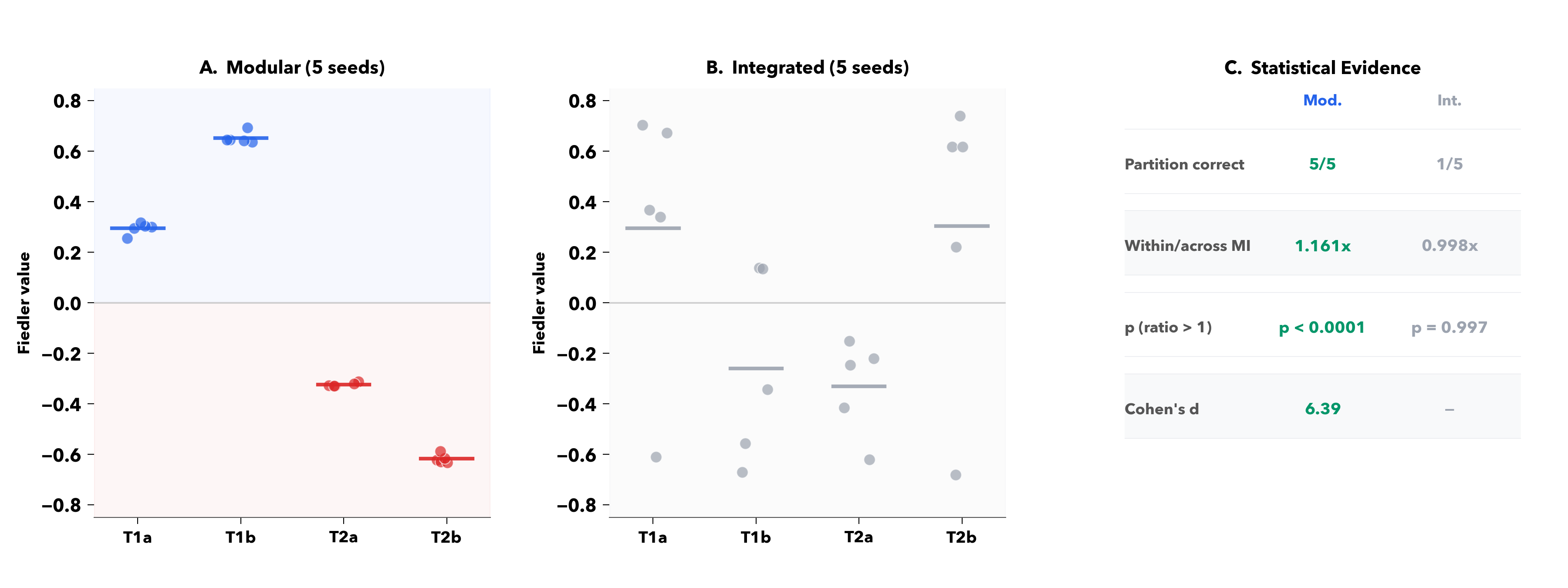}
\caption{Fiedler-vector values across five prompt seeds for the modular and integrated
conditions in Qwen3-0.6B.
\textbf{A.}~Modular condition: entities assigned to Team~1 (T1a, T1b; blue) consistently
receive positive Fiedler values, while Team~2 entities (T2a, T2b; red) receive negative
values. No overlap across any seed. Horizontal lines show per-entity means.
\textbf{B.}~Integrated condition: Fiedler values show no consistent grouping by role.
\textbf{C.}~Statistical summary.
All five modular seeds recover the planted partition; integrated matches by chance
(1/5). Paired comparison of the team-separation statistic $S$ across seeds is highly
asymmetric, with the modular distribution tightly concentrated near $S \approx 0.94$ and
the integrated distribution dispersed over [0,~0.84].}
\label{fig:fiedler_modular}
\end{figure}

In the modular condition, the Fiedler partition recovers the planted team assignment
$\{T1a,T1b\}$ versus $\{T2a,T2b\}$ in all five prompt seeds. In the integrated condition, the
partition is inconsistent, matching the team split in only one of five seeds (consistent with
chance for a bipartition of four items). Figure~\ref{fig:fiedler_modular}A shows the
distribution of Fiedler-vector values across seeds: in the modular condition, entities assigned
to Team~1 consistently receive positive values and entities assigned to Team~2 receive
negative values, with no overlap. In the integrated condition, no such pattern appears
(Figure~\ref{fig:fiedler_modular}B).

\begin{table}[h!]
\centering
\caption{Modular versus integrated conditions in Qwen3-0.6B (five prompt seeds,
200~prompts each, with name and slot-order permutation). $S$ is the Fiedler team-separation
statistic of Eq.~\eqref{eq:team_separation}; bracketed quantities are nonparametric bootstrap
95\,\% confidence intervals across seeds.}
\label{tab:modular}
\begin{tabular}{@{}lcc@{}}
\toprule
Metric & Modular & Integrated \\
\midrule
Partition correct                      & 5/5                       & 1/5 (chance) \\
Within/across ratio $R$                & 1.086                     & 1.004        \\
Team-separation $S$ (mean [95\,\% CI]) & 0.944 [0.938, 0.949]      & 0.303 [0.040, 0.598] \\
Modular $-$ Integrated $S$             & 0.641 [0.341, 0.906]      & --- \\
Paired $t$ on $S$ (mod.\ vs.\ int.)    & $t=3.97$, $p=0.017$       & --- \\
\bottomrule
\end{tabular}
\end{table}

The within/across ratio is modest ($R \approx 1.08$) but the partition itself is perfectly
consistent across all five seeds (Table~\ref{tab:modular}). The integrated condition
produces $R \approx 1.00$, confirming that the MI matrix is nearly uniform when no team
structure is described. The Fiedler team-separation statistic $S$ is tightly concentrated for
modular (0.94 with 95\,\% bootstrap CI [0.94, 0.95]) and broad for integrated (mean 0.30,
CI [0.04, 0.60]), and a paired $t$-test on $S$ across the five matched seeds confirms the
difference (Section~\ref{sec:results_modular_para}). The absolute MI ratio is small not
because the signal is weak but because the modular and integrated prompts induce broadly
similar levels of overall pairwise dependence; the diagnostic information lies in
\emph{which} pairs are coupled, which the partition captures and a scalar cross-MI measure
does not (Section~\ref{sec:results_crossmi}).

\label{sec:results_modular_para}

\subsection{LLM dynamic reassignment}
\label{sec:results_reassignment}

\begin{figure}[h!]
\centering
\includegraphics[width=\textwidth]{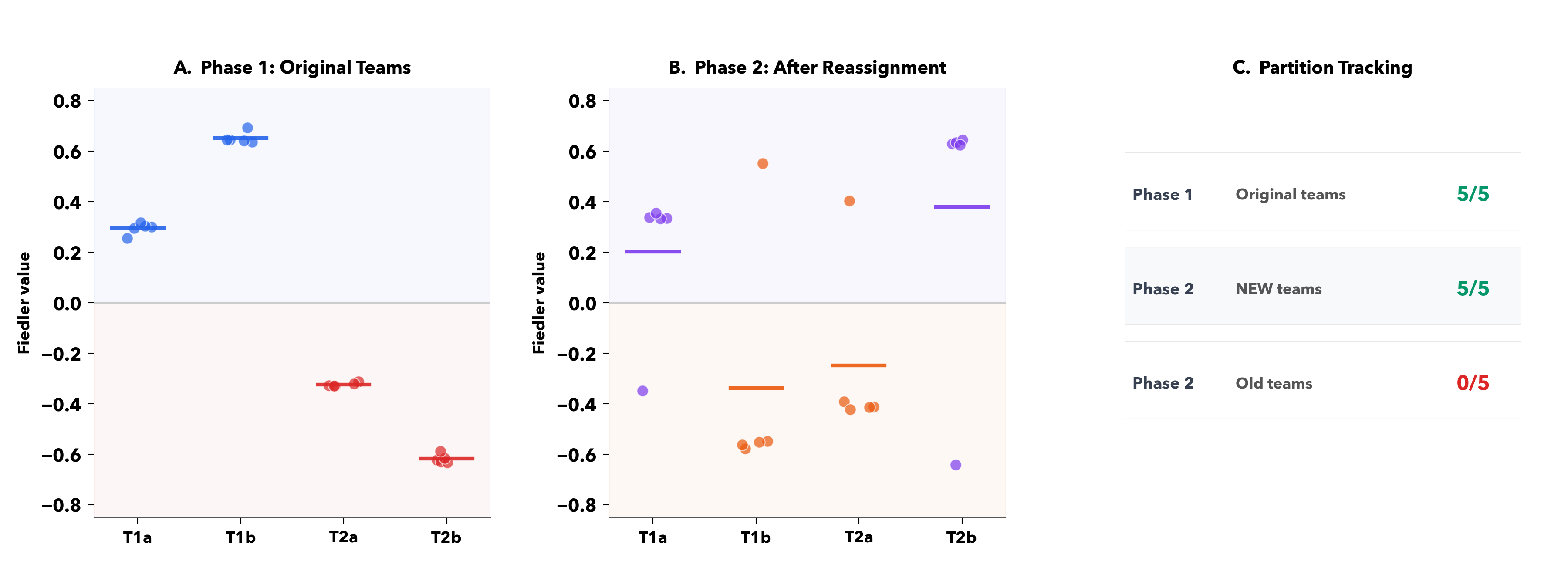}
\caption{Dynamic reassignment in Qwen3-0.6B.
\textbf{A.}~Phase~1 (original teams): Fiedler values separate Team~1 (T1a, T1b) from
Team~2 (T2a, T2b) across all five seeds.
\textbf{B.}~Phase~2 (after reassignment): Fiedler values now separate the \emph{new}
teams (T1a+T2b versus T1b+T2a). The partition tracks the described reassignment,
not the original structure.
\textbf{C.}~Summary: current-team partition correct in 5/5 seeds for both phases;
old-team partition matches 0/5 in Phase~2.}
\label{fig:fiedler_reassignment}
\end{figure}

When Phase~2 prompts describe a team reassignment (T1a now paired with T2b, T2a now paired
with T1b), the Fiedler partition tracks the new team structure in all five seeds and the
original structure in none (Table~\ref{tab:reassignment};
Figure~\ref{fig:fiedler_reassignment}). Comparing the team-separation statistic $S$ under
the new and old groupings of Phase~2 yields a highly significant paired difference
($S_{\mathrm{new}} = 0.972$ vs.\ $S_{\mathrm{old}} = 0.221$; $t = 94.3$, $p < 10^{-7}$).
This parallels the REINFORCE dynamic swap result (Section~\ref{sec:results_dynamic}): the
partition follows the described coalition structure, not a residual encoding of the original
assignment.

\begin{table}[h!]
\centering
\caption{Dynamic reassignment in Qwen3-0.6B (five prompt seeds). $S$ values are means of the
Fiedler team-separation statistic across the five seeds.}
\label{tab:reassignment}
\begin{tabular}{@{}lcc@{}}
\toprule
& Phase~1 & Phase~2 \\
\midrule
Partition matches current teams                          & 5/5  & 5/5  \\
Partition matches old teams                              & ---  & 0/5  \\
Team-separation $S$ under current teams                   & 0.944 & 0.972 \\
Team-separation $S$ under old teams                       & ---  & 0.221 \\
Paired $t$ on $S$ (current vs.\ old, Phase~2)             & ---  & $t=94.3$, $p<10^{-7}$ \\
\bottomrule
\end{tabular}
\end{table}

\subsection{LLM implicit coalitions}
\label{sec:results_implicit}

\begin{figure}[h!]
\centering
\includegraphics[width=\textwidth]{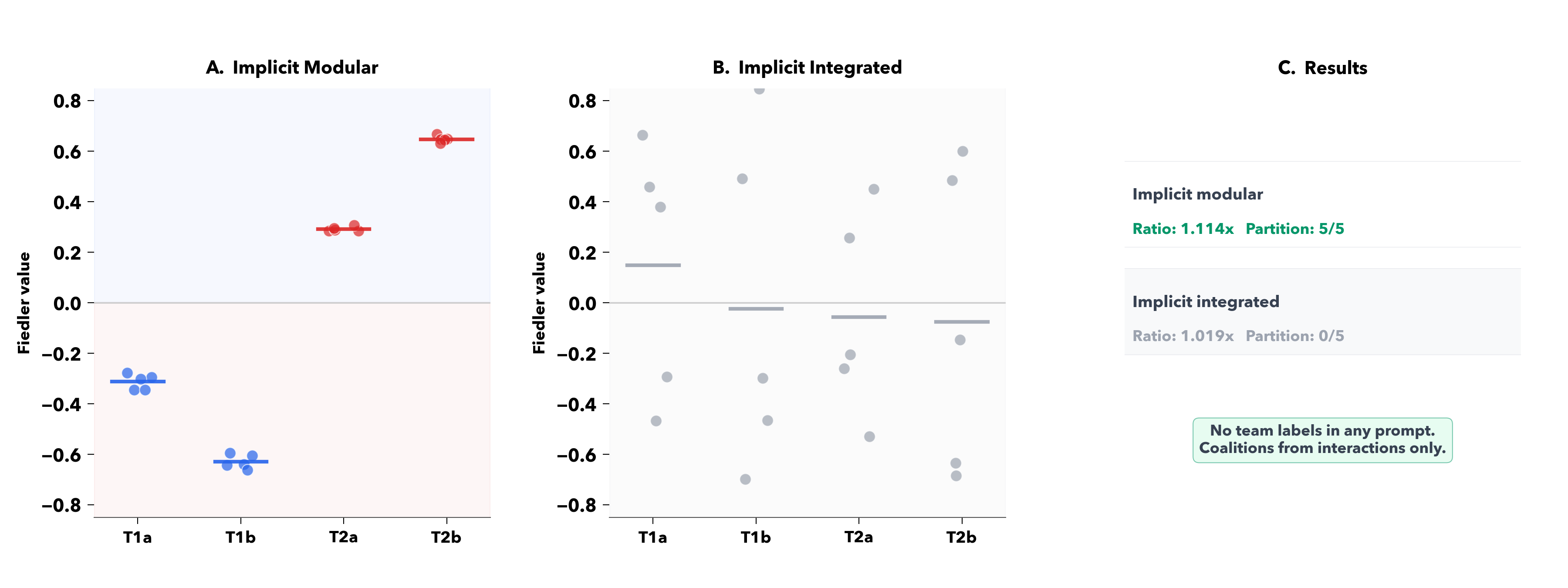}
\caption{Implicit coalition detection in Qwen3-0.6B: no team labels used.
\textbf{A.}~Implicit modular condition (interactions imply two teams): Fiedler values
separate the implied teams across all five seeds ($R = 1.075$, partition correct 5/5).
\textbf{B.}~Implicit integrated condition (all four entities interact equally): no
consistent separation ($R = 1.009$, partition correct 0/5).
\textbf{C.}~Summary. Coalition structure is recovered from described interaction
patterns alone, ruling out keyword matching on explicit team labels.}
\label{fig:fiedler_implicit}
\end{figure}

When team labels are removed entirely and coalition structure is conveyed only through
described interaction patterns (e.g., \emph{``Alice handed the report to Bob \ldots\
Separately, Carol shared data with Dave''}), the Fiedler partition still recovers the
implied two-team structure in all five seeds ($R = 1.075$;
Figure~\ref{fig:fiedler_implicit}). In the implicit integrated condition (all four entities
interact equally), the partition is inconsistent ($R = 1.009$, 0/5 seeds). A paired
comparison of the team-separation statistic $S$ between the two implicit conditions confirms
the dissociation: $S_{\mathrm{implicit\,modular}} = 0.941$ vs.\
$S_{\mathrm{implicit\,integrated}} = 0.253$, paired $t = 11.94$, $p = 2.8 \times 10^{-4}$.
This rules out the possibility that the modular partition is driven by keyword matching on
explicit team labels; the model's representations encode relational semantics inferred from
described interactions.

\subsection{LLM adversarial dissociation: labels versus interactions}
\label{sec:results_adversarial}

The adversarial experiment asks what the Fiedler partition tracks when team labels and
described interaction patterns conflict. Table~\ref{tab:adversarial} summarizes the results
across three sub-conditions, each replicated over five prompt seeds.

\begin{table}[h!]
\centering
\caption{Adversarial dissociation: what does the Fiedler partition track? Five prompt seeds
per condition, 200~prompts each. ``Label $S$'' is the team-separation statistic
\eqref{eq:team_separation} computed under the label-based partition $\{A,B\}$ vs.\
$\{C,D\}$; ``Interaction $S$'' is computed under the interaction-based partition $\{A,C\}$
vs.\ $\{B,D\}$. Paired $t$-tests compare the two within each condition.}
\label{tab:adversarial}
\begin{tabular}{@{}lccccc@{}}
\toprule
Condition & Partition tracks & Label $R$ & Interaction $R$ &
Label $S$ & Interaction $S$ \\
\midrule
Aligned (labels $=$ interactions)        & Labels (5/5)       & 1.16 & 0.98 & 0.94 & 0.05 \\
Dissociated (labels $\ne$ interactions)  & Labels (5/5)       & 1.15 & 0.98 & 0.96 & 0.03 \\
Interaction-only (no labels)             & Interactions (5/5) & 0.93 & 1.19 & 0.05 & 0.99 \\
\bottomrule
\end{tabular}
\end{table}

The dissociation is striking. In the aligned and dissociated conditions, the
team-separation statistic under the label partition is 0.94 and 0.96 respectively, while
under the interaction partition it is essentially zero (0.05 and 0.03). In the
interaction-only condition the statistics flip almost perfectly: label $S$ collapses to
0.05 and interaction $S$ rises to 0.99. Within each condition the paired $t$-test
comparing the two candidate partitions is significant at $p < 10^{-5}$
(aligned: $t=138.5$, $p=1.6\times 10^{-8}$;
dissociated: $t=132.8$, $p=1.9\times 10^{-8}$;
interaction-only: $t=-48.2$, $p=1.1\times 10^{-6}$).

\begin{figure}[h!]
\centering
\includegraphics[width=\textwidth]{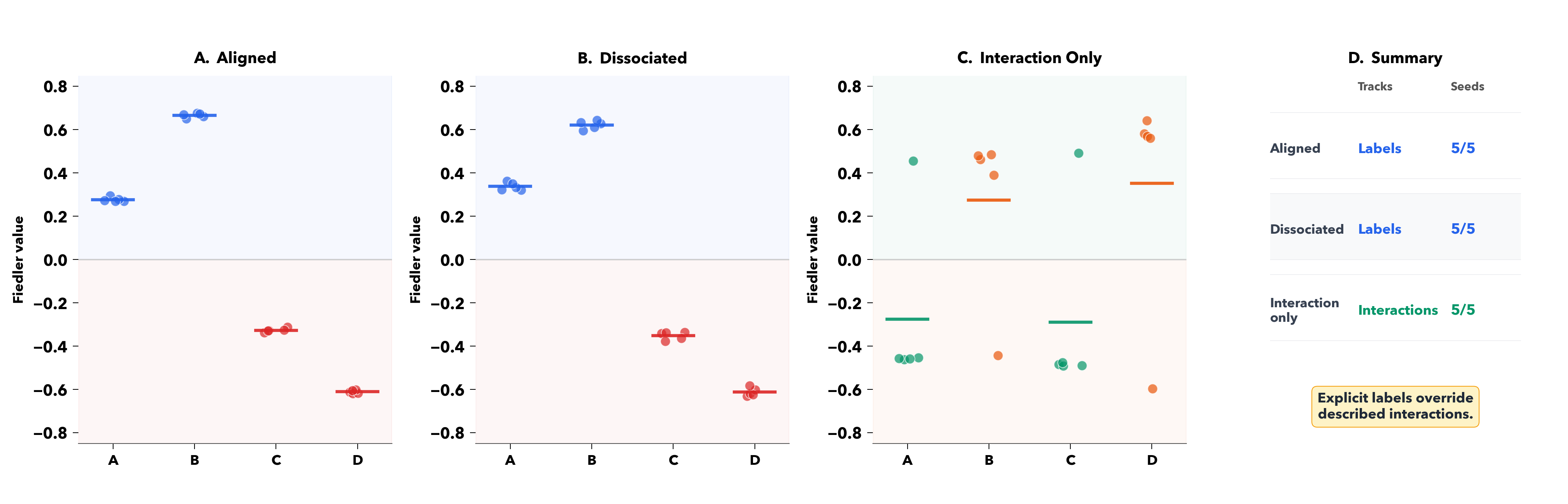}
\caption{Adversarial dissociation: labels versus interactions.
\textbf{A.}~Aligned condition: Fiedler values separate entities by label-based teams
(blue above zero, red below), consistent across all five seeds.
\textbf{B.}~Dissociated condition: despite described interactions conflicting with
labels, the partition still tracks label-based teams (5/5).
\textbf{C.}~Interaction-only condition: with labels removed, the partition flips to
track the described interaction pattern---entities that interact (green) separate from
those that are isolated (orange), 5/5 seeds, $R = 1.19$.
\textbf{D.}~Summary. Explicit team labels override described interaction patterns in
the model's representations; interaction structure emerges only when labels are absent.}
\label{fig:fiedler_adversarial}
\end{figure}

When explicit team labels are present---even when they directly contradict the described
interaction pattern---the partition tracks the \emph{label} structure
(Figure~\ref{fig:fiedler_adversarial}A--B). When labels are removed, the partition flips to
track the \emph{interaction} structure (Figure~\ref{fig:fiedler_adversarial}C). The
interaction-only condition produces the strongest within/across ratio observed in any LLM
experiment ($R=1.19$), suggesting that labels act as a competing signal that partially
masks the interaction-based MI structure.

This finding has a practical implication for alignment monitoring: when analyzing LLM
representations for coalition structure, explicit relational framing in the input can dominate
over the relational patterns actually described. Removing or controlling for explicit labels
may be necessary to expose the underlying interaction-based representational organization.

\subsection{Comparison to scalar cross-agent mutual information}
\label{sec:results_crossmi}

A natural baseline question is whether the partition information recovered by
$\Phi_{\mathrm{spectral}}$ could equally be obtained from a simpler scalar measure of
representational coupling, in particular the total cross-agent mutual information
$T(M) = \sum_{i<j} M_{ij}$. To test this we compute $T(M)$ alongside the within/across
ratios under the candidate partitions described in
Sections~\ref{sec:results_modular}--\ref{sec:results_adversarial}.

The clearest case is the adversarial dissociation experiment, where the same four entities
appear under three different relational framings. Table~\ref{tab:crossmi_adv} reports the
total cross-MI for each adversarial condition together with the within/across ratios under
both candidate partitions.

\begin{table}[h!]
\centering
\caption{Total cross-agent mutual information versus structural ratios in the adversarial
dissociation experiment. The total $T(M)$ is essentially constant across the three
conditions (coefficient of variation $\approx$ 4\,\%), but the within/across ratios under
the two candidate partitions reorganize completely. A scalar cross-MI measure cannot
distinguish which partition is informationally privileged; the Fiedler partition can.}
\label{tab:crossmi_adv}
\begin{tabular}{@{}lccc@{}}
\toprule
Condition & Total $T(M)$ & Label-partition $R$ & Interaction-partition $R$ \\
\midrule
Aligned                & 3.59 & 1.16 & 0.98 \\
Dissociated            & 3.49 & 1.15 & 0.98 \\
Interaction-only       & 3.81 & 0.93 & 1.19 \\
\bottomrule
\end{tabular}
\end{table}

In all three conditions the four entities exhibit comparable total pairwise
representational coupling. A statistic of the form $T(M)$ would treat them as
indistinguishable. The Fiedler bipartition, by contrast, returns qualitatively different
coalition structures across conditions: in aligned and dissociated prompts, the partition
is the label split $\{A,B\}\,|\,\{C,D\}$; in the interaction-only prompts, it flips to
the interaction split $\{A,C\}\,|\,\{B,D\}$ despite the total dependence being slightly
\emph{higher} than in the other conditions.

A similar qualitative point applies to the modular versus integrated comparison
(Section~\ref{sec:results_modular}). Here the two conditions do differ in total cross-MI
(modular $T(M) = 2.58$, integrated $T(M) = 1.13$), but this difference reflects an overall
shift in pairwise dependence induced by the more verbose modular prompts and is by itself
ambiguous: it does not say whether the additional dependence is concentrated within
candidate teams or spread uniformly across the four entities. The within/across ratio
$R(A^\star,B^\star)$ resolves the ambiguity (modular 1.16 vs.\ integrated 1.00), but only
because the partition $(A^\star,B^\star)$ is supplied by the spectral cut. Replacing
$\Phi_{\mathrm{spectral}}$ with a scalar baseline therefore loses precisely the structural
information---the membership list---that makes the method useful for monitoring and
oversight (cf.\ Section~2.4).

In short, $\Phi_{\mathrm{spectral}}$ contributes to coalition detection something a scalar
cross-MI measure cannot: it identifies \emph{which} agents are coupled, not merely
\emph{how much} coupling is present in the system as a whole. This complements the
behavioral-versus-representational dissociation established by the negative control
(Section~\ref{sec:results_negative}) and is what enables the partition flip observed in
Section~\ref{sec:results_adversarial}.

\section{Discussion}
\label{sec:discussion}

The main contribution of this paper is to recast $\Phi_{\mathrm{spectral}}$ as a tool for coalition detection rather than only as a whole-system integration statistic. In the present use case, the partition itself is primary. A coalition is operationally defined as a subset of agents whose hidden states are more tightly coupled to one another than to the rest of the system, and the Fiedler bipartition provides a scalable way to recover that boundary from observed representations alone. Read this way, the results support a strong but limited claim: representational mutual-information structure can reveal coalition boundaries that are absent, ambiguous, or misleading at the behavioral level \cite{BaileySchneider2026phi,Mediano2022WeakIIT}.

\paragraph{Behavioral coordination versus representational coupling.}
The negative control in Section~\ref{sec:results_negative} is especially important for interpreting the method. Agents independently trained to match the same group oracle produce near-perfect within-group behavioral agreement, yet the mutual-information graph built from independent inputs contains no recoverable coalition structure. This dissociation shows that the method is not merely clustering agents by similar outputs. Instead, it responds when agents' internal states statistically encode common partners, shared local contingencies, or other forms of representational co-adaptation. For safety and alignment, that distinction is critically important. Many benign systems will display behavioral similarity without hidden coalition formation, while some of the most consequential coalition shifts may first appear in internal organization before they become obvious in aggregate behavior.

\paragraph{Understanding the dynamic results.}
The dynamic experiments extend this point from static recovery to online monitoring. In the REINFORCE swap setting (Section~\ref{sec:results_dynamic}), the post-swap partition follows new reward-defined partners even after the transient behavioral disruption has washed out. In the LLM reassignment setting (Section~\ref{sec:results_reassignment}), mid-layer token representations reorganize when the described team structure changes. Together, these results suggest that the method is useful not only for identifying coalitions retrospectively, but also for tracking coalition reorganization as representational dependencies evolve.

\paragraph{What the LLM bridge does and does not show.}
The LLM experiments are best interpreted as a bridge between literal multi-agent interaction and representational modularity in foundation models. The implicit-coalition condition (Section~\ref{sec:results_implicit}) shows that explicit team labels are not required: described interaction patterns alone can induce modular structure in hidden-state space. At the same time, the adversarial condition (Section~\ref{sec:results_adversarial}) reveals that explicit labels dominate interaction descriptions when both are present. This is substantively interesting, but it also exposes a methodological caution. In language models, the recovered partition reflects the model's representational organization under a particular prompt distribution, not an observer-independent ``true'' coalition structure. Prompt framing is therefore part of the measured system. The token-positions-as-agents design should be understood as evidence that the spectral method transfers to foundation-model representations, not as proof that a single language model contains multi-agent coalitions in the same sense as interacting reinforcement-learning agents.

\paragraph{Relation to integrated information and weak-IIT framing.}
These findings fit naturally within the weaker, observer-relative interpretation of $\Phi_{\mathrm{spectral}}$ developed by Bailey and Schneider \cite{BaileySchneider2026phi} and the broader weak-IIT program \cite{Mediano2022WeakIIT}. The present measure is pairwise, undirected, and statistical; it is not an estimate of exact causal $\Phi$ in the sense of IIT \cite{Tononi2004InformationIntegrationTheory,Oizumi2014IIT30,Albantakis2023IIT40}. For coalition detection, however, this limitation is also an advantage. The question at issue is practical decomposability of observed representational structure: do the hidden states of these agents behave as one nearly uniform block, or do they fall into relatively weakly coupled subgroups? The Fiedler partition provides a tractable answer to that question without requiring full causal reconstruction. In this sense, the method is better viewed as an observer-level diagnostic of modular organization than as a doctrinal test of intrinsic system unity.

This framing also clarifies why the integrated condition is a useful control rather than a failure case. When the MI graph is close to uniform, many cuts are nearly equivalent and the Fiedler vector is weakly constrained \cite{Luxburg2007SpectralClustering,BaileySchneider2026phi}. In a whole-system integration setting, that degeneracy limits interpretability. In the present coalition-detection setting, however, near-uniformity is exactly what one would expect when no nontrivial subgroup structure is present. The integrated prompts therefore provide a desirable null: the method should \emph{not} force a stable partition when all four entities are represented as a single cohesive team.

\paragraph{Limitations.}
Several limitations follow directly from the current design. First, the method compresses high-dimensional relationships into pairwise MI edges, so higher-order synergy or redundancy may be missed \cite{Mediano2019MeasuringIntegratedInformation}. Second, MI is observational rather than interventional: it cannot by itself distinguish direct interaction from common input, architectural bias, or prompt-induced structure \cite{Barrett2011PracticalMeasures}. The independent-input negative control shows one way to manage this problem, but comparable controls will be needed in other settings. Third, the LLM effects, although highly consistent, are numerically modest in absolute terms; the strong paired statistics in Section~\ref{sec:results_modular} should be read as reflecting low across-seed variance in the Fiedler structure rather than a large absolute separation in the underlying mutual-information matrix. Fourth, the bridge experiments use a single relatively small model, one extraction layer, four entity tokens, and templated prompt families. Larger models, different layers, multi-turn dialogues, and genuinely interactive LLM populations may exhibit qualitatively different representational geometry.

A further limitation concerns ontology. In the REINFORCE setting, the nodes of the graph correspond to distinct learning agents with explicit reward coupling. In the LLM setting, the nodes correspond to token positions within one forward pass. The same mathematics applies in both cases, but the interpretation does not. The bridge between them is useful, yet it should not erase the difference between distributed social interaction and structured single-model representation.

\paragraph{Future directions.}
These limitations point to a clear research agenda. On the methodological side, it will be important to compare alternative MI estimators, quantify partition stability, and extend the framework to directed or perturbational statistics that better distinguish causal influence from common drive. On the empirical side, the most natural next steps are larger LLMs, richer prompt ecologies, longer interaction horizons, and genuinely multi-agent language-model settings in which agents exchange messages, develop conventions, or strategically conceal coalition formation \cite{Ashery2025EmergentSocialConventions,Rahwan2019MachineBehaviour}. Recursive decomposition is also likely to become more valuable at larger population sizes, where hidden coalitions may be nested, overlapping, or transient rather than flat. For alignment and oversight, the most plausible near-term role of the method is as a screening tool: a way to flag emerging representational coalitions for deeper causal or behavioral inspection, rather than as a standalone detector of collusion or intent.

\section{Conclusions}
\label{sec:conclusions}

This paper introduced $\Phi_{\mathrm{spectral}}$ as a practical method for detecting coalition structure from internal neural representations. Across controlled REINFORCE experiments and bridge experiments in Qwen3-0.6B, the method recovered hierarchical grouping, tracked reassignment, rejected a behavioral false positive, and revealed a substantive dissociation between label-based and interaction-based organization. The central lesson is that coalition structure is often more legible in hidden-state dependence than in overt behavior alone.

The contribution is therefore both empirical and methodological. Empirically, the results show that internally coupled subgroups can be detected even when behavioral monitoring is silent, transient, or actively misleading. Methodologically, they show that a simple MI-graph plus Fiedler-bipartition pipeline can serve as a scalable observer-relative diagnostic of modular organization in distributed AI systems. At the same time, the method does not by itself establish causal coupling, collusive intent, or intrinsic integration in the strong IIT sense. Its output should be interpreted as evidence of representational organization under the measurement protocol used.

Under that interpretation, $\Phi_{\mathrm{spectral}}$ offers a promising addition to the alignment and oversight toolbox. As multi-agent and population-level AI systems become more capable, the ability to detect hidden coalitions from internal representations may become as important as monitoring overt performance. With larger-scale validation, stronger causal controls, and extensions to richer multi-agent settings, the present approach could become a useful component of the broader toolkit for monitoring emergent organization in advanced AI systems.

\section*{Acknowledgments}

This work was initiated while C.B. was Research Director at AE Studio. We thank AE Studio for supporting the early stages of this research.

\printbibliography

\end{document}